\documentclass[sigconf,]{acmart}

\AtBeginDocument{%
  }

\copyrightyear{2025}
\acmYear{2025}
\setcopyright{cc}
\setcctype{by-nc}
\acmConference[MAD'25]{4th ACM International Workshop on Multimedia AI against Disinformation }{June 30-July 3, 2025}{Chicago, IL, USA}
\acmBooktitle{4th ACM International Workshop on Multimedia AI against Disinformation (MAD'25), June 30-July 3, 2025, Chicago, IL, USA}
\acmDOI{10.1145/3733567.3735573}
\acmISBN{979-8-4007-1891-5/2025/06}

\begin{document}

\title{SocialDF: Benchmark Dataset and Detection Model for Mitigating Harmful Deepfake Content on Social Media Platforms}

%
\author{Arnesh Batra}

\orcid{0009-0002-0086-7870}
\affiliation{%
  \institution{Indraprastha Institute of Information Technology Delhi}
  \city{New Delhi}
  \state{Delhi}
  \country{India}
}
\email{arnesh23129@iiitd.ac.in}

\author{Anushk Kumar}
\authornote{Equal Contribution.}
\orcid{0009-0002-4752-5077}
\affiliation{%
  \institution{Indraprastha Institute of Information Technology Delhi}
  \city{New Delhi}
  \state{Delhi}
  \country{India}
}
\email{anushk23115@iiitd.ac.in}

\author{Jashn Khemani}
\orcid{0009-0003-5128-1631}
\authornotemark[1] 
\affiliation{%
  \institution{Indraprastha Institute of Information Technology Delhi}
  \city{New Delhi}
  \state{Delhi}
  \country{India}
}
\email{jashn23256@iiitd.ac.in}

\author{Arush Gumber}
\orcid{0009-0003-8448-5390}
\authornotemark[1] 

\affiliation{%
  \institution{Indraprastha Institute of Information Technology Delhi}
  \city{New Delhi}
  \state{Delhi}
  \country{India}
}
\email{arush23136@iiitd.ac.in}

\author{Arhan Jain}
\orcid{0009-0009-7656-2345}
\affiliation{%
  \institution{Indraprastha Institute of Information Technology Delhi}
  \city{New Delhi}
  \state{Delhi}
  \country{India}
}
\email{arhan23118@iiitd.ac.in}

\author{Somil Gupta}
\orcid{0009-0009-1532-2196}
\affiliation{%
  \institution{Indraprastha Institute of Information Technology Delhi}
  \city{New Delhi}
  \state{Delhi}
  \country{India}
}
\email{somil24559@iiitd.ac.in}



\begin{abstract}
The rapid advancement of deep generative models has significantly improved the realism of synthetic media, presenting both opportunities and security challenges. While deepfake technology has valuable applications in entertainment and accessibility, it has emerged as a potent vector for misinformation campaigns, particularly on social media. Existing detection frameworks struggle to distinguish between benign and adversarially generated deepfakes engineered to manipulate public perception.

To address this challenge, we introduce SocialDF, a curated dataset reflecting real-world deepfake challenges on social media platforms. This dataset encompasses high-fidelity deepfakes sourced from various online ecosystems, ensuring broad coverage of manipulative techniques. We propose a novel LLM-based multi-factor detection approach that combines facial recognition, automated speech transcription, and a multi-agent LLM pipeline to cross-verify audio-visual cues. Our methodology emphasizes robust, multi-modal verification techniques that incorporate linguistic, behavioral, and contextual analysis to effectively discern synthetic media from authentic content.
\end{abstract}






\begin{teaserfigure}
    \centering
    \includegraphics[width=\textwidth]{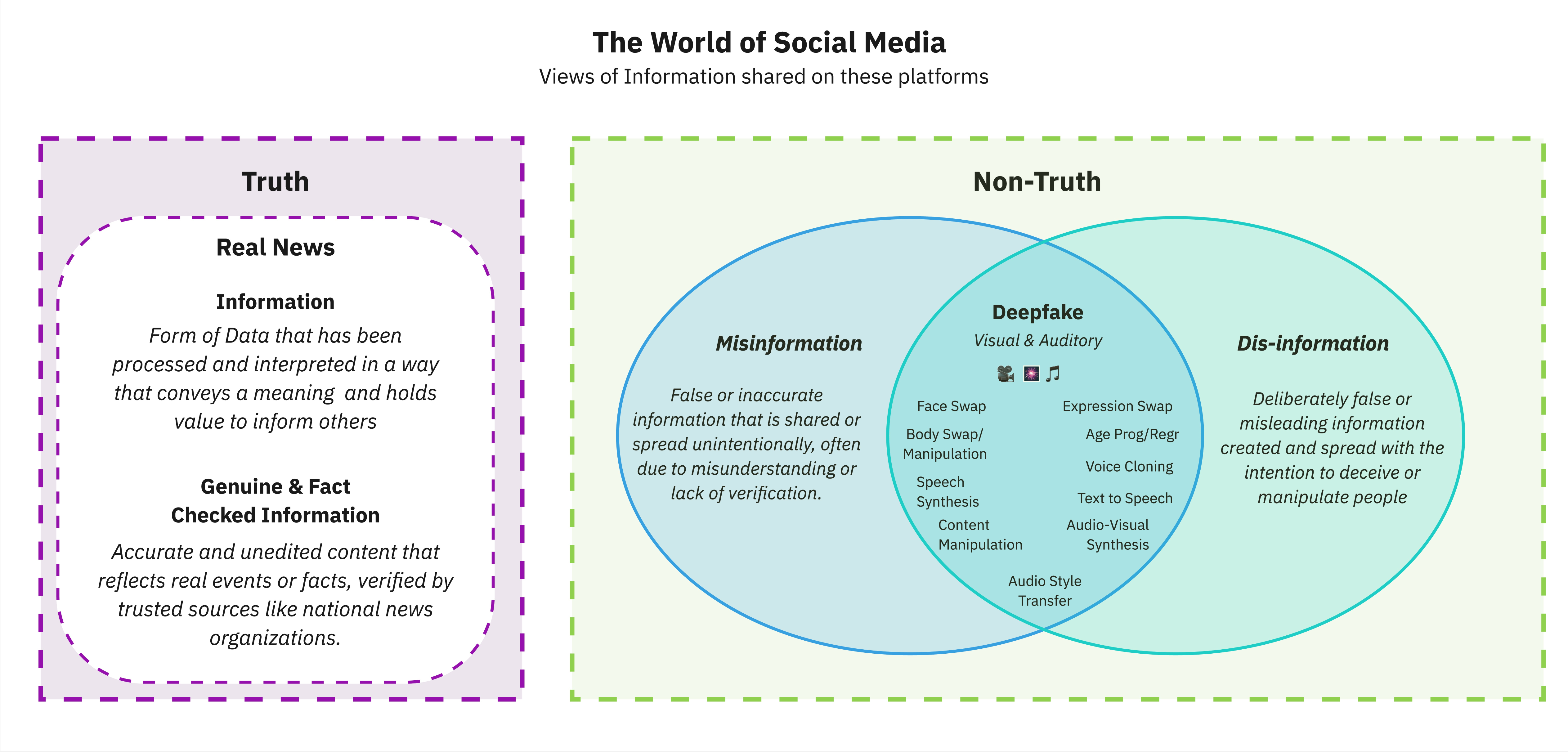}
    \caption{A visual breakdown of information on social media, categorizing it into Truth (real, fact-checked news) and Non-Truth (misinformation, disinformation, and deepfakes), it highlights the role of deepfake techniques like face swaps and voice cloning in spreading manipulated content. The figure also reflects definitions of misinformation, disinformation and malinformation as outlined in EU Code of Practice (2022) \cite{CodeOfPractice}.}
    \label{fig:information}
\end{teaserfigure}

\maketitle
\begin{figure*}[t]
    \centering
    \includegraphics[width=0.8\textwidth]{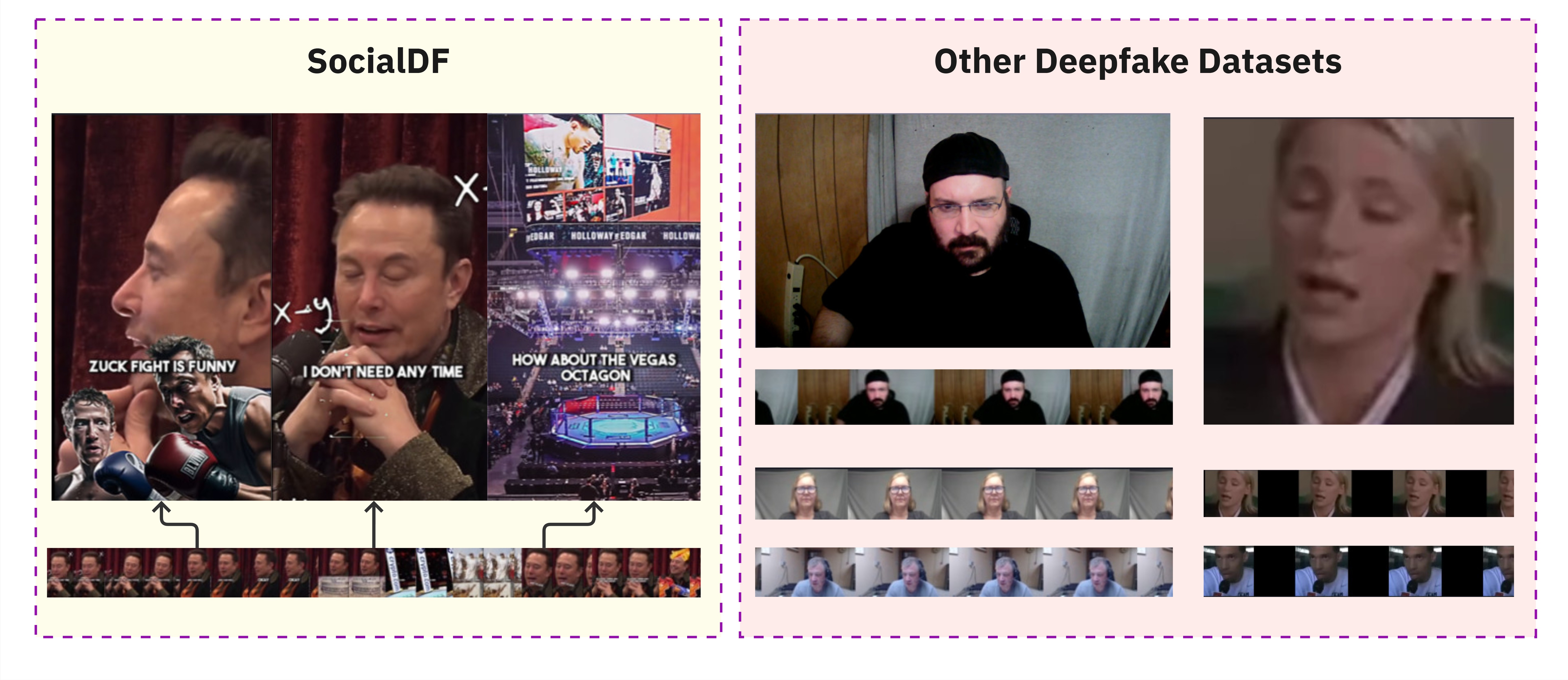}
\caption{
Comparison between our SocialDF dataset (left) and other deepfake datasets (right). While existing datasets show clear, single-speaker videos in controlled settings, SocialDF includes overlays, scene changes, and multiple speakers—making it more representative of real-world social media content for robust model evaluation.}

    \label{fig:information}
\end{figure*}

\section{Introduction}
Deepfakes have rapidly transformed digital media by merging advanced machine learning with accessible content creation tools. While initially celebrated for creative applications, deepfakes now pose serious risks by blurring the line between authentic and fabricated content. Today, even non-experts can produce convincing deepfakes that mimic public figures, fueling misinformation.
\begin{table*}[h]
    \centering
    \caption{Comparison of Audio-Video Deepfake Detection Datasets}
    \begin{tabular}{|l|c|c|l|}
        \hline
        \textbf{Dataset} & \textbf{Real Samples} & \textbf{Fake Samples} & \textbf{Analysis} \\
        \hline
        FakeAVCeleb \cite{khalid2022fakeavcelebnovelaudiovideomultimodal} &  570 & 25,000 & Generated using publically available Softwares; Low Quality \\

        \hline
        LAV-DF \cite{cai2022you} & 36,431 &  99,873 & Generated using publically available Softwares; Low Quality \\
        \hline

        \hline
        AV-Deepfake1M \cite{cai2024avdeepfake1mlargescalellmdrivenaudiovisual} & 500,000 & 500,000 & Generated using publically available Softwares; Low Quality \\
        \hline

        \hline
        DeepSpeak v1.0 \cite{barrington2024deepspeakdatasetv10} &  6,226 &  6,226 & Good Quality; Less variety and camera angles \\
        \hline
        
        \textbf{SocialDF (Ours)} & \textbf{1,071} & \textbf{1,055} & \textbf{Realistic; real-world deepfakes; Very high quality} \\
        \hline
    \end{tabular}
    \label{tab:deepfake_datasets}
\end{table*}
Social media platforms have emerged as the primary battleground for deepfake proliferation. The ease of sharing and rapidly consuming short-form videos enables actors to distribute manipulated content that can alter public perceptions and erode societal trust. Incidents involving fabricated speeches, manipulated endorsements, and impersonated public figures demonstrate the profound impact of deepfakes on discourse and security. Moreover, the dynamic and noisy environment of social media challenges traditional detection methods that rely solely on visual or temporal inconsistencies.

The societal implications of deepfake technology extend beyond misinformation. Hostile entities exploit these tools to incite discord, undermine democratic processes, and compromise privacy. A 2023 study by Chemerys \cite{inproceedings} illustrates this threat, documenting a cyber-incident during the Russian-Ukrainian conflict where threat actors disseminated a fabricated video of President Zelenskyy simulating a surrender declaration.
\newline
Our contributions in this work are threefold:

We provide an in-depth analysis of the current deepfake landscape, exploring both its creative potential and its risks for misinformation and societal discord.
We introduce a novel, context-aware deepfake detection framework that integrates multi-modal data and leverages state-of-the-art machine learning techniques to improve detection accuracy and resilience.
We evaluate our approach using a diverse dataset that mirrors real-world scenarios, demonstrating the framework's scalability and effectiveness 
The remainder of this paper is organized as follows. Section 2 surveys related work in deepfake generation and detection, highlighting key challenges and opportunities. Section 3 details our dataset collection methodology and discusses its use Section 4 presents the techniques we use to develop our fact checking framework. Section 5 presents experimental results and a comparative analysis with existing techniques., and Section 6 concludes with directions for future research.

By addressing these critical issues, our research aims to contribute to the development of more secure and transparent digital media ecosystems, ensuring that technological innovation is harnessed responsibly and ethically.

\subsection{Proposed Approach}
To tackle the challenges of deepfake misinformation, we propose:  
\begin{itemize}
    \item \textbf{SocialDF} – A benchmarking dataset comprising 2,126 deepfake and real videos sourced from social media, capturing state-of-the-art manipulations.  
    \item \textbf{Fact Checking Framework} – A novel architecture integrating multimodal analysis to detect and mitigate deceptive video content.  
\end{itemize}  
These contributions aim to enhance deepfake detection and combat misinformation effectively.

\section{Related Work}
There has been a mass increase in the amount of deepfake videos, which has led to many methods to target such videos. Primarily, there are two methods that humans use to differentiate deepfakes from regular videos; the first way is to use video-audio features to check for artifacts or lipsyncing, and the second one is to see what the deepfake video is portraying. However, not all individuals\cite{doi:10.1177/2327857924131023} can effectively assess these aspects, as most deepfakes pertain to specific domains where domain-specific knowledge significantly enhances one’s ability to recognize such content.

\subsection{Existing Datasets}

Deepfake detection datasets can generally be categorized into three types: those containing visual samples, audio samples, and multimodal datasets that include both audio and video. Audio-only datasets, while useful for detecting synthetic speech, lack crucial contextual cues such as the speaker's identity and the visual alignment of facial expressions with speech. Conversely, visual-only datasets struggle to capture conversational context, making it difficult to assess inconsistencies in speech dynamics, such as unnatural prosody or mismatched lip movements.

Multimodal datasets \cite{barrington2024deepspeakdatasetv10} \cite{DFDC2020} \cite{celebdf} \cite{cai2023reallymeanthatcontent} \cite{cai2024avdeepfake1mlargescalellmdrivenaudiovisual}, which integrate both audio and visual modalities, are considered the most robust for deepfake detection, as they enable cross-modal verification through audio-visual synchronization analysis, speech-lip consistency checks, and facial expression tracking. However, a significant limitation of existing state-of-the-art deepfake datasets is their oversimplified nature—most samples depict subjects with clear, unobstructed faces, speaking directly to the camera under controlled conditions. This controlled setting makes it relatively easy to identify fakes using straightforward audio-visual features, such as lipsync accuracy and facial blending artifacts.

In real-world scenarios, deepfakes are often more sophisticated, featuring occlusions, side profiles, background noise, varied lighting conditions, cut scene changes, multiple people, and adversarial manipulations designed to evade detection systems. We aim to bridge the gap by presenting a deepfake and fact-checking dataset - SocialDF.

\subsection{Fact Checking}
In recent years, various methods have been proposed for detecting deepfakes, with lip-sync/audio based approaches \cite{sun2024diffusionfakeenhancinggeneralizationdeepfake} \cite{liu2024lips} \cite{khalid2022fakeavcelebnovelaudiovideomultimodal} \cite{katamneni2024contextualcrossmodalattentionaudiovisual} being one of the most explored techniques. These methods primarily focus on analyzing the alignment between the audio and facial movements, detecting discrepancies that could indicate manipulation. However, lip-sync approaches are limited in scenarios involving scene changes, multiple individuals, or when the person is not consistently visible throughout the video. In such cases, lip-sync-based methods face challenges in maintaining accuracy, as the altered facial expressions or lip movements cannot be reliably matched to the audio. Furthermore, deepfakes often involve complex manipulations where the individual’s identity or appearance is changed, or periods of obscurity make lip synchronization ineffective. In contrast, fact/misinformation checking methods provide a more comprehensive detection strategy. By analyzing the broader context of the video, including the consistency of the narrative with established facts, these methods can detect discrepancies that go beyond facial analysis. This makes fact-checking a more robust approach in addressing the challenges posed by deepfakes, particularly in situations where lip-syncing alone would fail to identify manipulation.
Existing fact-checking research has primarily focused on images and text \cite{singal-etal-2024-evidence} \cite{van2023detectingcorrectinghatespeech}. In this work, we propose an enhanced architecture that extends and improves upon these approaches, making them suitable for the video domain.
\begin{figure*}[t]
    \centering
    \includegraphics[width=0.8\textwidth]{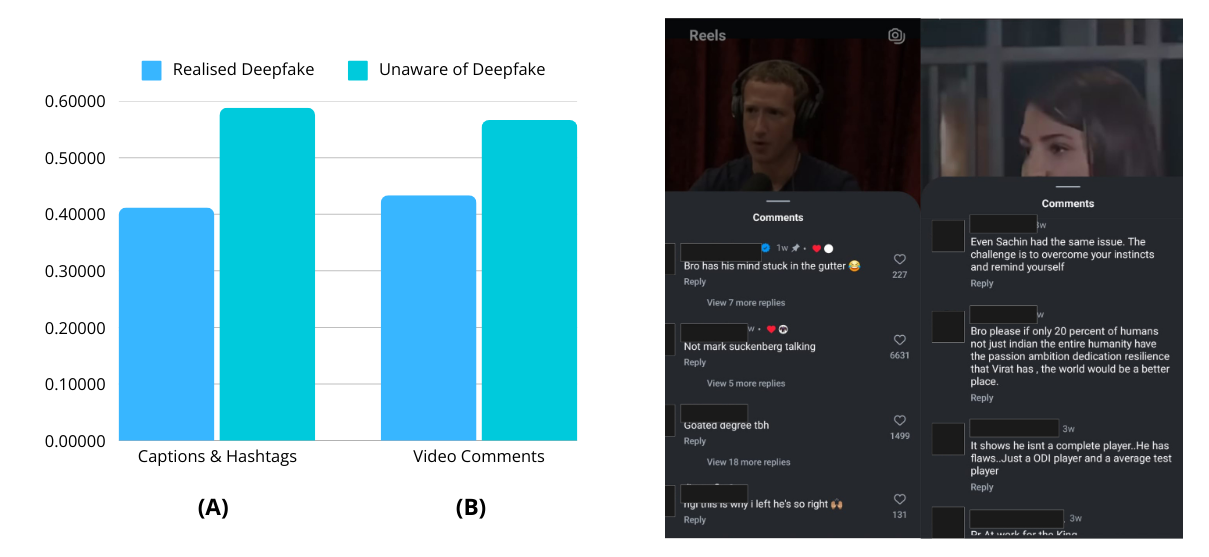}
    \caption{Table showcasing the distribution of users who identified the video as deepfake based on comments and mentions in caption or hashtag by the author. Screenshot from deepfake videos highlighting user comments, revealing the inability of many viewers to differentiate between real and fake content.}
    \label{fig:comments}
\end{figure*}
\section{Dataset}
Dataset Description and Significance:
SocialDF comprises 869 potential deepfake targets and 2,126 short-form videos (1,071 genuine, 1,055 manipulated). The targets represent popular figures from professions most susceptible to deepfakes, primarily featuring celebrities and influential personalities. We sourced content from social media platforms with rapid-consumption formats like Reels and Stories, where users often view content without critical scrutiny. This environment provides an ideal context to study real-world deceptive content that is difficult to identify through casual viewing.


Data Collection Process:
To compile this dataset, we employed both manual and automated approaches. We created a list of popular personalities which are potential targets as deepfakes, running an automated process to get up to 20 images of each personality. On the manual side, annotators used keyword-based searches (e.g., “deepfake” ,“face swap” ,"parody" or the names of specific celebrities known to be frequent targets of manipulation) to locate potential deepfake videos. We also scoured popular Instagram accounts reputed for posting face manipulations sometimes comedic, sometimes malicious so as to encompass diverse content. We prioritized videos that were especially challenging to classify by the naked eye, aiming to capture borderline cases that even experienced viewers might mistake for genuine footage. This manual selection was supplemented by automated routines to systematically scrape posts from relevant hashtags or user profiles.
\begin{figure}[t]
    \centering
    \includegraphics[width=1\linewidth]{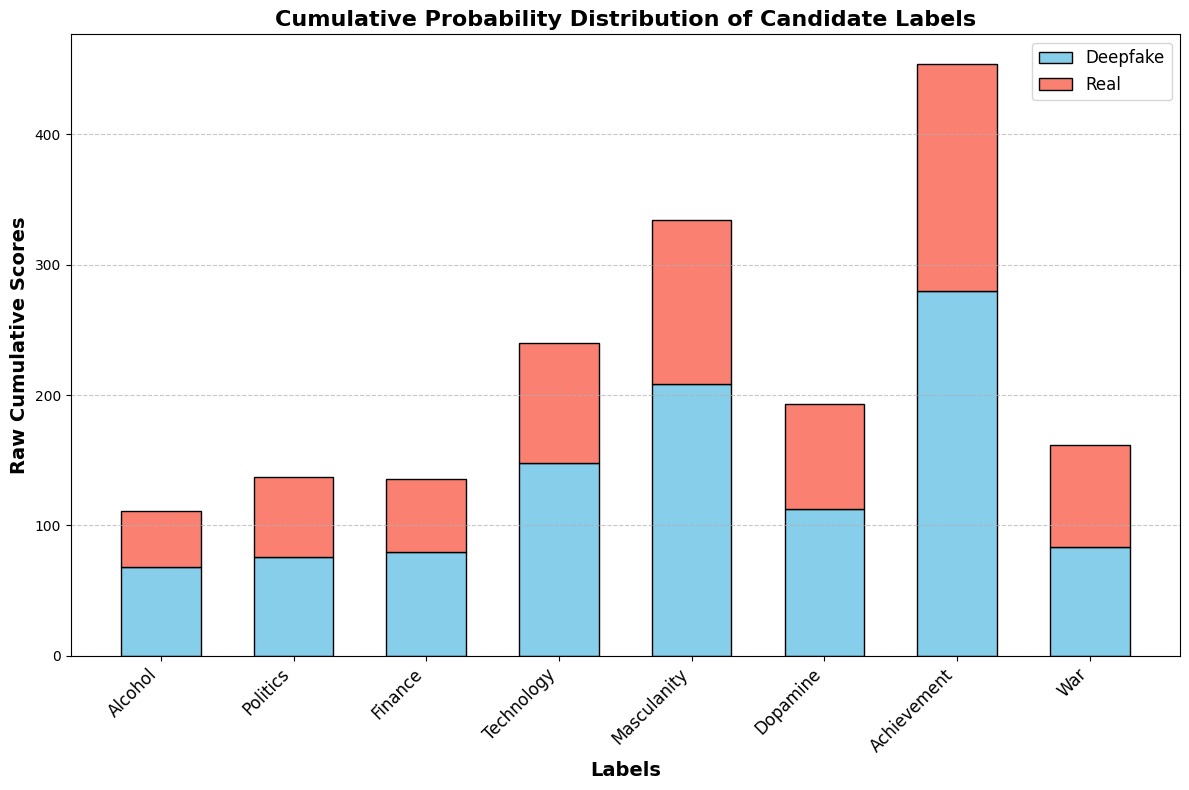}
    \caption{Cumulative count of category labels for Deepfake and Real samples across different topics. The stacked bar chart displays raw cumulative scores for each label, highlighting the distributions among Deepfake (blue) and Real (red) content.}
    \label{fig:graph}
\end{figure}

To refine the labels (real/fake) beyond our initial suspicion, we relied on uploader-provided cues such as hashtags like \texttt{\#deepfake}, mentions of tools like Parrot AI, or captions explicitly stating the video was generated or altered. These signals were treated as primary indicators of fake content. Real videos were collected from credible or verified accounts with no mention of synthetic content. In ambiguous cases, we performed manual review of comments using zero-shot classification to support labeling, followed by consensus-based verification where needed. While our annotators were not professional fact-checkers, this combination of uploader intent and cross-verified comments helped maintain high label reliability without introducing subjective bias.

Data Analysis and Characteristics:
We performed a large-scale sentiment analysis on the collected comments to discern user perceptions of authenticity. As shown in Figure~\ref{fig:comments} (B), the distribution of sentiment scores ranging from strong agreement with the content’s authenticity to skepticism or outright accusations of fakery. The inability of viewers to differentiate between real and fake content is evident. These insights provide context on how often and how quickly real-world audiences recognize manipulated content. Surprisingly, preliminary results suggest that a sizeable fraction of viewers fail to spot deepfakes, reaffirming the pressing need for reliable detection methods.

\begin{figure*}[t]
    \centering
    \includegraphics[width=\textwidth]{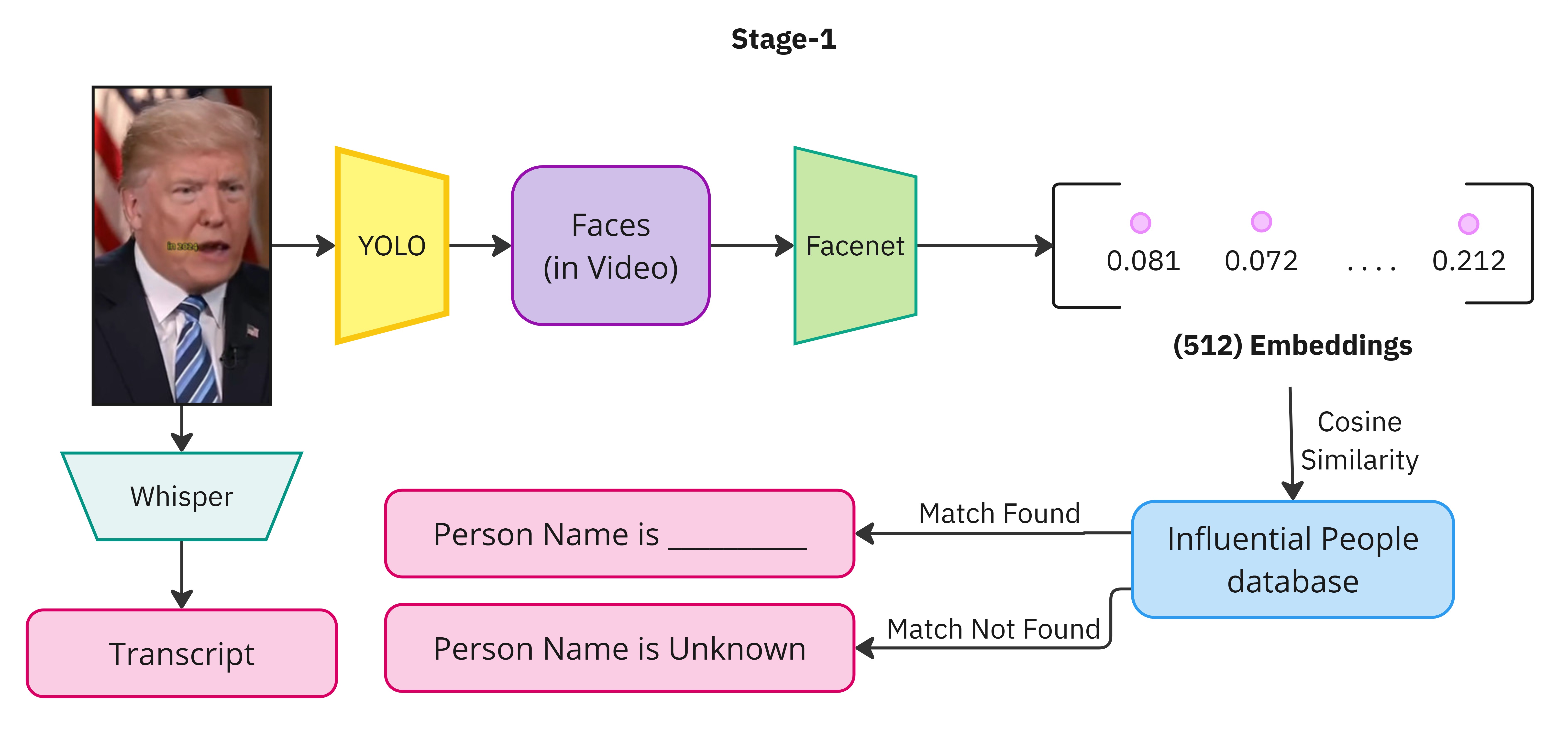}
    \caption{Stage 1: Identifying which persons are in the video and what they are speaking to get context of the conversation.}
    \label{fig:stage1}
\end{figure*}


Genre and Person-of-Interest Distribution:
Our dataset covers a range of genres—political speeches, music videos, comedic sketches, and promotional clips—to ensure broad coverage of real-world contexts where deepfakes appear.

Our Instagram based dataset offers a realistic distribution of manipulated content, unlike existing datasets that rely on staged or controlled samples. This authenticity enhances the generalizability of detection models to real-world scenarios.

In addition to a balanced set of real and fake videos, the dataset includes rich contextual metadata such as user comments, sentiment scores, and popularity indicators—supporting the development of robust, context-aware detection systems. By blending manual and automated verification across diverse genres, SocialDF provides a practical resource to bridge the gap between experimental methods and the complex realities of social media.




\subsection{Potential Uses of our Dataset}

\paragraph{Audio Deepfake Detection}  
One of the primary applications of the dataset is in the detection of audio deepfakes, the audio in this dataset is very difficult to be determined as fake by an average human being.

\paragraph{Audio-Visual Deepfake Detection}  
Given the growing sophistication of deepfake technologies, detecting deepfakes that involve both audio and video components has become increasingly important.

\paragraph{Social Media Analysis}  
Social media platforms are hotspots for the spread of misinformation, often in the form of deepfakes or manipulated content. Our dataset can be used to monitor and analyze content on social media, helping to identify and flag potential deepfakes or harmful media. 

\paragraph{Multimodal Fact-Checking}  
The dataset can also be applied to multimodal fact-checking systems, where both text and multimedia content (such as audio and video) are examined for accuracy.

\paragraph{Potential Victims}
The dataset can be used to specifically identify deepfakes of famous individuals, safeguarding the reputation of potential victims and avoid miscommunication among the viewers.

\section{Fact Checking Framework}
We propose a fact-checking approach for detecting deepfake videos designed to spread misinformation, particularly those targeting specific individuals. These videos constitute the majority of deepfake content circulating on social media. Our approach consists of a two-step pipeline for detecting video falsification. In the first stage, we identify the individuals present in the video and transcribe their spoken content. This step is accomplished through a combination of face recognition and automatic speech recognition (ASR) techniques, ensuring accurate speaker identification and transcription.
In the second stage, we leverage the extracted identity and speech information in conjunction with a Large Language Model (LLM) \cite{naveed2024comprehensiveoverviewlargelanguage} to assess the authenticity of the video. The LLM processes these inputs to analyze inconsistencies, contextual anomalies, and semantic deviations, ultimately computing the probability that the video has been manipulated. This probabilistic assessment serves as a reliable indicator of potential falsification.
By integrating multimodal analysis—visual recognition, speech transcription, and language-based reasoning—our approach enhances the robustness of deepfake detection, improving the reliability of authenticity verification in digital media.
\subsection{1st Stage}
The first step is determining whether there are any influential people in the video and identifying them. The face recognition process is initiated by analyzing video frames to detect and recognize human faces. In the first step, the video is processed frame by frame, where each frame undergoes detection and localization of faces using YOLO (You Only Look Once) \cite{redmon2016lookonceunifiedrealtime}, an efficient object detection model. YOLO's ability to detect multiple objects in real-time makes it suitable for face detection within dynamic video environments.

Once a face is detected, the region of interest (ROI) containing the face is cropped and passed through FaceNet \cite{Schroff_2015}, a deep learning-based facial feature extraction model. FaceNet generates a 512-dimensional embedding vector for each detected face. This embedding is a unique numerical representation of the person's facial features, capturing the intrinsic characteristics of the face in a high-dimensional space. FaceNet's embedding vector is crucial for differentiating between individuals, even in cases of subtle variations in facial expressions, lighting conditions, or angles.

The generated facial embeddings are then compared against a pre-existing database of influential people containing 869 people, using cosine similarity \cite{Steck_2024} to determine whether a match exists. Cosine similarity is employed to measure the angular distance between the embeddings, which quantifies how similar two faces are based on their vector representations. 
The cosine similarity between two vectors \( \mathbf{A} \) and \( \mathbf{B} \) is given by the following formula:

\[
\text{Cosine Similarity} = \frac{\mathbf{A} \cdot \mathbf{B}}{\|\mathbf{A}\| \|\mathbf{B}\|}
\]

Where:

- \( \mathbf{A} \) and \( \mathbf{B} \) are the embedding vectors for two faces. \( \mathbf{A} \cdot \mathbf{B} \) represents the dot product of the two vectors. \( \|\mathbf{A}\| \) and \( \|\mathbf{B}\| \) are the magnitudes (Euclidean norms) of the vectors.
\begin{figure*}[t]
    \centering
    \includegraphics[width=0.8\textwidth]{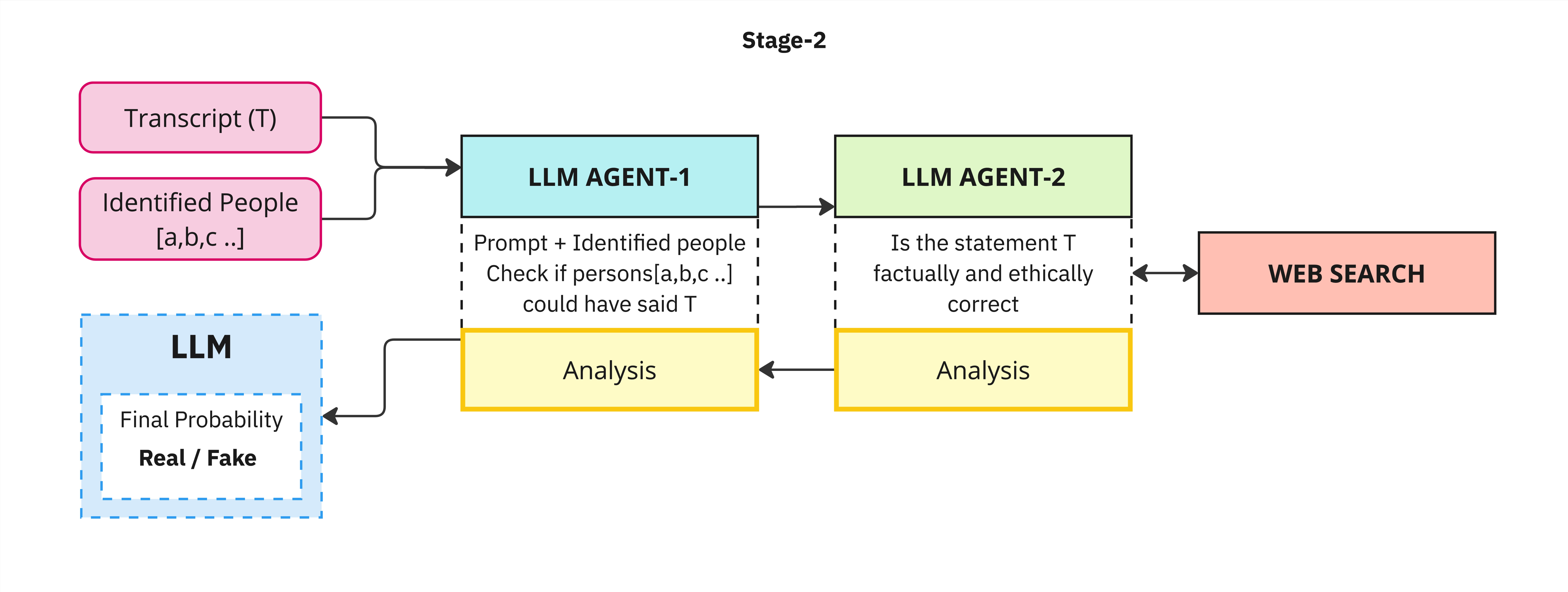}
    \caption{Stage 2: Multi-agent pipeline for accurately detecting fake information spreading videos.}
    \label{fig:stage2}
\end{figure*}
Simultaneously, the system processes the audio from the video using Whisper \cite{radford2022robustspeechrecognitionlargescale}, an automatic speech recognition (ASR) model. Whisper transcribes the spoken content into text, providing a structured transcript of what was said in the video. 

At the end of this stage, the system produces two key outputs:
\begin{itemize}
\item \textbf{Identified Individuals} – The names of all individuals identified in the video through facial recognition, or "Unknown" if no match is found for a particular individual.

    \item \textbf{Transcript} – The complete text of spoken content derived from the video’s audio.
\end{itemize}

\subsection{2nd Stage}
The proposed architecture utilizes a multi-agent pipeline based on Large Language Models (LLMs) to detect deepfakes by rigorously analyzing the authenticity and ethical validity of textual transcripts. This system is designed to assess whether a given statement, attributed to specific individuals, aligns with their known patterns of communication and is both factually accurate and ethically sound. Each LLM agent within the pipeline is equipped with access to a web search tool, enabling real-time retrieval of external information to enhance the reliability and context-awareness of their evaluations. By integrating both authenticity verification and ethical analysis, the architecture establishes a robust framework for combating misinformation propagated through deepfake content.

Each LLM agent within the pipeline is equipped with access to a web search tool, enabling real-time retrieval of external information to enhance the reliability and context-awareness of their evaluations. However, we ensure that this retrieval does not leak clues from the test sample itself — videos under evaluation are short-form clips rarely indexed or ranked high in web results, and our pipeline does not use metadata like titles or descriptions. Instead, the LLM assesses only the transcript and identity match to determine whether the spoken content aligns with what the individual could plausibly say or whether it is factually accurate.

    The input to the system consists of a transcript ($T$) and a list of identified people ($[a, b, c, \dots]$) who are purported to have made the statement(s) in $T$.

\textbf{LLM Agent-1:} This module receives the transcript ($T$) along with the identified people as input.  
A prompt is constructed using $T$ and the list of individuals ($[a, b, c, \dots]$). The prompt is designed to query whether the identified individuals could plausibly have made the statements in $T$.  
The output is a detailed analysis indicating the plausibility of attribution based on contextual, stylistic, and semantic alignment.

\textbf{LLM Agent-2:} This module evaluates the factual correctness and ethical implications of the statements in $T$.  
It leverages web search to retrieve supporting evidence or counterexamples for the claims in $T$.  
A secondary analysis assesses the ethical considerations, ensuring that the statements do not propagate misinformation, harmful content, or ethical violations.

The final LLM module consolidates the analyses performed by the initial two agents, incorporating both their outcomes and the underlying reasoning behind their evaluations. Based on this comprehensive synthesis, the final module determines whether the video content is authentic or a deepfake. Since the system uses only the transcript and identified individuals as input, and short-form social media videos rarely appear in top-ranked web results, there is no risk of inadvertently retrieving metadata such as video titles or descriptions during web search. This integrative, multimodal approach enhances the accuracy and reliability of the system in distinguishing real content from fabricated material.

\section{Experiments and Results}
We initially evaluated our dataset using LipFD \cite{liu2024lips}, the current state-of-the-art (SOTA) lip-sync detection model. LipFD was trained from scratch on our SocialDF dataset using a 90/10 train-test split. The split was stratified to maintain equal proportions of real and fake videos across both sets, with no overlap in video clips or subjects.
 Based on its performance on our dataset, we subsequently developed and refined our proposed framework. Our experiments reveal that LipFD plateaus at 51.24\% accuracy on our dataset (as shown in Table \ref{tab:epoch_stats}). The likely reason is that LipFD assumes continuous, close-up footage of a speaker’s face, focusing heavily on lip-sync consistency. In contrast, our real-world dataset is replete with cut-scenes, multiple people speaking, and heavy on-screen text or graphics poses significant challenges and LipFD struggles to maintain reliable lip-tracking and consistently misclassifies or defaults to predicting the “real” class. These observations underscore the inherent mismatch between models trained on tightly cropped “talking-head” benchmarks and the actual complexity of short-form videos on social media platforms. As a result, purely lip-centric approaches rapidly degrade in the presence of occlusions, scene changes, speech overlap, and re-edited clips, justifying a shift toward richer, multimodal fact-checking models such as ours.

\begin{table}[h]
\centering
\begin{tabular}{|c|c|c|c|c|}
\hline
\textbf{Epoch} & \textbf{Training Loss} & \textbf{Val Acc. (\%)} & \textbf{FPR (\%)} & \textbf{FNR (\%)} \\ \hline
0              & 62.07                  & 51.24                             & 0.00             & 99.19             \\ \hline
1              & 57.80                  & 51.04                             & 0.00             & 99.60             \\ \hline
3              & 56.34                  & 51.34                             & 0.00             & 98.99             \\ \hline
5              & 55.67                  & 50.94                             & 0.00             & 99.80             \\ \hline
10             & 55.83                  & 51.04                             & 0.00             & 99.60             \\ \hline
15             & 55.72                  & 51.64                             & 0.00             & 98.38             \\ \hline
20             & 60.18                  & 50.94                             & 0.00             & 99.80             \\ \hline
25             & 55.95                  & 51.04                             & 0.00             & 99.60             \\ \hline
30             & 55.71                  & 51.14                             & 0.00             & 99.39             \\ \hline
35             & 57.90                  & 50.94                             & 0.00             & 99.80             \\ \hline
\end{tabular}
\caption{Epoch-wise statistics for LipFD on our dataset, highlighting the plateau in validation accuracy and high false negative rates.}
\label{tab:epoch_stats}
\end{table}

The LipFD model's training plateaus at around 50\%–51\% accuracy within the first 5–10 epochs, demonstrating its difficulty adapting to our real-world dataset. Despite a steadily decreasing training loss (e.g., ~62.07 at epoch 0 to ~57.70 at epoch 10), the validation loss remains stagnant, and the accuracy sees negligible improvement. Key statistics include a false negative rate (FNR) consistently near 99\% and a false positive rate (FPR) of 0\%, indicating a strong bias toward predicting the "real" class while failing to generalize across cut-scenes, occlusions, and speech overlaps. This highlights the model's reliance on clean, tightly cropped training data, which fails to translate to noisy, multimodal environments.


While LipFD excels at exploiting temporal inconsistencies in lip motions and audio, its inability to perform well on our dataset highlights significant limitations: it relies heavily on consistently visible and close-up lip regions, which makes it ineffective in real-world scenarios involving partial occlusions, overlapping text, or frequent scene cuts. Moreover, it struggles with complex multispeaker scenarios, where rapid transitions and multiple active speakers disrupt its assumptions of a single, consistently tracked face. Although the method introduces perturbation handling (e.g., noise, blur), it lacks robustness against editorial-style changes typical of social media content. Future directions could include integrating multimodal fact-checking and contextual learning to better handle dynamic, multilingual, and noisy environments.
In sum, LipFD contributes a noteworthy method for lip-sync forgery detection, but its narrow focus on “clean” single-speaker data imposes significant constraints in complex, real-world scenarios. Our empirical results demonstrate that a broader fact-checking pipeline is better suited to handle the multifaceted nature of deepfake content on social media.
\newline
Seeing the performance of LipFD, we tested our novel method on SocialDF, which uses a multimodal approach for fact-checking. For the Large Language Models (LLMs) in our framework, we experimented with several open-source options, including Llama 3.3 \cite{llama3modelcard}, Qwen \cite{qwen2.5-1m}, and the DeepSeek R-1 \cite{deepseekai2025deepseekr1incentivizingreasoningcapability} reasoning model. We experimented with temperature values of 0.3, 0.5, and 0.7 for the LLMs. A temperature of 0.5 yielded the best accuracy, offering a balanced trade-off between deterministic outputs (as seen with 0.3) and creative variability (as seen with 0.7), allowing the model to reason effectively without hallucinating.
Through extensive testing, we found that the optimal results were achieved with a temperature value of 0.5, striking a balance between diversity and determinism in the model's outputs. 
For the web search, we utilized the DuckDuckGo search engine due to its rapid performance and commitment to a no-ads policy, which ensures an efficient and uninterrupted search experience. For video transcription, we employed Whisper Large V3 Turbo \cite{radford2022whisper}, a model recognized for its exceptional speed and near state-of-the-art accuracy. This model achieves a significant reduction in transcription time by optimizing its architecture, specifically by decreasing the number of decoder layers from 32 to 4, resulting in a model that is approximately six times faster than its predecessors with minimal loss in accuracy.

Among the tested models, our framework demonstrated the highest accuracy and reliability when paired with DeepSeek R-1, which consistently outperformed its counterparts across various metrics.

The framework's performance was evaluated using two key metrics: Accuracy and F1-Score. These metrics were computed for each tested Large Language Model (LLM) to determine the most effective model for deepfake detection. Our experiments showed that DeepSeek R-1 provided the best overall results, achieving the highest Accuracy and F1-Score. Qwen and Llama 3.3 also demonstrated competitive performance but fell short compared to DeepSeek R-1.

The following table summarizes the performance of the tested LLMs:

\begin{table}[h!]
\centering
\begin{tabular}{|l|c|c|}
\hline
\textbf{Model}         & \textbf{Accuracy (\%)} & \textbf{F1-Score} 


\\ \hline
Llama 3.3 8B             & 89.5                   & 0.90              \\ \hline
Qwen 2.5 7B                  & 87.4                   & 0.89              \\ \hline
\textbf{DeepSeek R-1 Llama 8B} & \textbf{90.4}          & \textbf{0.93}     \\ \hline
\end{tabular}

\label{tab:results}
\end{table}

From the results we can see that DeepSeek R-1 Llama 8B excels in detecting misinformation-spreading deepfakes due to its advanced reasoning. The model is trained using large-scale reinforcement learning (RL) and chain of thought mechanism, which enhances its ability to perform complex reasoning tasks such as self-verification and reflection. These skills are crucial for identifying and analyzing the nuanced patterns often present in deepfake content.

\section{Future Work}
The proposed dataset has the potential to significantly enhance existing LipSync and Audio Deep Fake Detection models or contribute to the development of innovative solutions in this domain. As the quality of DeepFake technology continues to improve and achieve greater realism over time, the dataset can be extended to reflect these advancements, enabling models to stay up-to-date and ensure accurate, efficient detection.

To strengthen the benchmark's robustness, future work will also include adversarial resistance testing, where existing and proposed models are evaluated against adversarially perturbed videos or those specifically crafted to bypass detection. This will help assess the real-world resilience of detection systems.

Moreover, we plan to benchmark additional state-of-the-art deepfake detection models on SocialDF beyond LipFD, enabling a more comprehensive and comparative evaluation across detection paradigms.

Furthermore, governments, corporate entities, and social media platforms can leverage state-of-the-art detection models to strengthen content verification processes, automatically classify content based on severity, and take appropriate action against malicious creators. These models can also be optimized and integrated into mobile applications to facilitate real-time fact-checking of content, including videos featuring individuals, thus promoting trust and accountability in digital media.

\section{Conclusion}
    In this research study, we presented a novel dataset and method to determine the authenticity of Deepfake / Synthetically Generated Content spreading over social media platforms, which, when reached the masses, could spread hatred and conflicts among them. In this emerging society, the accompanying Technologies of Artificial Intelligence and generative content  emphasize the need for robust detection frameworks. The data collection for the dataset is being sourced from sources that are more accessible to the general public, and there is scope for sharing this kind of content. During the data collection phase,  A significant group of participants faced difficulties in determining the authenticity of media content, often confused.   These contents are often the major promoters of debates and conflicts among varied people. Addressing these challenges, the proposed framework and dataset offer a robust solution to mitigate misinformation, enhance content verification, and promote informed decision-making, thereby contributing to the development of a more resilient and ethically driven AI-powered society.

\section{Ethical Statement}
Our dataset comprises short-form videos collected from publicly accessible social media platforms such as Instagram Reels and Stories. These videos were either uploaded by users themselves or are publicly shared content that explicitly or implicitly signals the use of generative or manipulated media. To respect copyright and platform terms of service, we provide only references (e.g., URLs) to original sources and release only annotations and metadata for research purposes. No copyrighted or non-consensual content is redistributed.

The dataset may contain manipulated media generated using publicly available deepfake tools. We relied on uploader disclosures (hashtags like \#deepfake, mentions of tools such as Parrot AI, or captions describing manipulation) and user-generated comments to label videos. All content flagged as manipulated was cross-verified through consensus review to ensure high label reliability. Our methodology is aligned with practices from prior work in multimedia research and falls under fair use as outlined in U.S. Code (2023), particularly for research, commentary, and educational purposes.

We also acknowledge ethical concerns about amplifying harmful or deceptive content. To address this, we took deliberate steps to include a balanced dataset with both real and fake videos and provide contextual cues like uploader intent, audience sentiment, and platform engagement. Videos were selected not to sensationalize but to reflect borderline, real-world scenarios where distinguishing manipulated from genuine media is inherently difficult.

Bias is another consideration. Our dataset may reflect demographic skew due to platform trends (e.g., more male personalities or English-speaking content). We document this explicitly and encourage researchers to treat these biases as important experimental variables. Furthermore, since the dataset involves public figures, we ensured content did not involve private individuals or violate expectations of privacy.

Finally, we emphasize that the dataset is intended solely for academic research, including the development of detection models, fact-checking tools, and misinformation awareness. No component of this dataset should be used for impersonation, harassment, or content generation purposes. The dataset will be made available under a CC BY-NC 4.0 license, ensuring use only for non-commercial, ethically sound purposes.

\section{Limitations}
Our dataset is primarily composed of short-form videos sourced from social media, which introduces certain limitations. Most notably, the focus on celebrities and high-profile individuals may restrict the generalization of detection models to less-public figures or everyday users. Additionally, relying on social media as the primary data source introduces biases tied to platform-specific trends, content styles, and temporal shifts—factors that can impact the long-term relevance and completeness of the dataset as online behavior continues to evolve. Furthermore, we observe that visual signals alone are often insufficient for accurate deepfake detection; audio content plays a critical role in identifying inconsistencies such as speech mismatches or identity violations, making multimodal analysis essential.

\section{Data and Code Availability}
The dataset and code used in this study are publicly available at the following GitHub repository: \url{https://github.com/arnesh2212/SocialDF/tree/main}. 

\bibliographystyle{ACM-Reference-Format}
\bibliography{sample-base}


\end{document}